\documentclass{article}

\usepackage[dblblindworkshop,final]{neurips_2026}

\usepackage[utf8]{inputenc} %
\usepackage[T1]{fontenc}    %
\usepackage{hyperref}       %
\usepackage{url}            %
\usepackage{booktabs}       %
\usepackage{amsfonts}       %
\usepackage{amssymb}
\usepackage{nicefrac}       %
\usepackage{microtype}      %
\usepackage{xcolor}         %
\usepackage{enumitem}
\usepackage{physics}
\usepackage{cleveref}       %
\usepackage{tikz}           %
\usepackage{wrapfig}
\usepackage{todonotes}

\workshoptitle{Representations for the Physical Sciences Workshop}
\title{
  Euclidean Fourier Neural Operators
}

\author{%
  Nathanael Bosch\thanks{Equal contribution.}
  \quad
  Niklas Frederik Schmitz\footnotemark[1]
  \quad
  Michael F.~Herbst
  \\
  Mathematics for Materials Modelling (MatMat)\\
  Institute of Mathematics \& Institute of Materials\\
  École Polytechnique Fédérale de Lausanne, 1015 Lausanne, Switzerland \\
  \texttt{\{nathanael.bosch, niklas.schmitz, michael.herbst\}@epfl.ch} \\
}

\begin{document}

\maketitle

\begin{abstract}
  Fourier neural operators (FNOs) provide an efficient framework for learning mappings between function spaces
  as they are, by construction, independent of the grid resolution at which they are trained and evaluated.
  However, FNOs are not independent of the periodic domain they are applied to:
  their discrete spectral weights are indexed by integer Fourier mode numbers,
  which correspond to physical wavevectors.
  When applied to a different domain, the same trained weights act at different wavevectors, and the FNO silently represents a different operator.
  This makes FNOs unsuitable for tasks where transfer across domains is crucial.
  We propose Euclidean Fourier neural operators~(EFNOs) as a domain-independent alternative to FNOs.
  By parameterizing the spectral kernel as a continuous function of the physical wavevector, the EFNO can learn operators that act consistently across periodic domains of varying shape and size.
  We evaluate the EFNO on a simple heat equation and on a practically relevant materials science task of learning exchange-correlation potentials across different crystal structures,
  and demonstrate that the EFNO is able to generalize to unseen grid sizes and domains.
\end{abstract}

\section{Introduction}
\label{sec:intro}
Fourier neural operators (FNOs) \citep{li2020fourier} have been established as an efficient framework for learning mappings between function spaces,
with successful applications throughout the physical sciences, including
fluid mechanics \citep{li2022fnoles},
weather and climate forecasting \citep{pathak2022fourcastnet, bonev2023sfno},
and seismology \citep{yang2021seismic}.
A particular strength of FNOs
is that they are, by construction, independent of the grid resolution at which they are trained and evaluated.
But many physical problems are not tied to a single fixed domain,
and the same problem can be posed on domains of varying shape and size.
This requires models that can transfer across domains.

FNOs do not natively support such transfer.
Their defining \emph{spectral convolution} layer parameterizes a convolution kernel by discrete trainable weights indexed by Fourier mode numbers~$m \in \mathbb{Z}^d$,
which correspond to physical wavevectors $k_m \in \mathbb{R}^d$ that depend on the periodic domain (see \cref{sec:fno}).
Thus, when a trained FNO is applied to a different domain, the same weights act at different wavevectors and the FNO silently represents a different operator.
Various extensions to FNOs have been proposed,
for applications to irregular grids or varying geometries
\citep{li2022geo, liu2023dafno, oguadimma2026geometry}
or via continuous parameterizations of the spectral kernel
\citep{xiao2024amortized, shi2026sirenfno, shin2024pdno, lee2026kano},
but their spectral layers still operate on Fourier mode indices and thus retain the same domain dependence.

In this paper, we propose \emph{Euclidean Fourier neural operators} (EFNOs).
By parameterizing the spectral kernel as a continuous function of the physical wavevectors, EFNOs learn operators in Euclidean space that can be applied consistently across domains.
We evaluate the EFNO on a simple heat equation and on
a materials science task that covers crystal structures of varying shape and size,
and demonstrate that, in contrast to the FNO, the EFNO is able to learn operators that generalize to unseen grid sizes and domains.

\section{Fourier neural operators and their domain dependence}
\label{sec:fno}

Neural operators are a class of neural networks designed to learn mappings between function spaces.
The common blueprint to parameterize a neural operator
$\mathcal{G}_\theta: \mathcal{A} \to \mathcal{U}$
uses a sequence of layers
\begin{align}
  \mathcal{G}_\theta := \mathcal{Q} \circ \sigma(\mathcal{W}_T + \mathcal{K}_T) \circ \dots \circ \sigma(\mathcal{W}_1 + \mathcal{K}_1) \circ \mathcal{P},
\end{align}
with point-wise linear lifting layer $\mathcal{P}$ and projection layer $\mathcal{Q}$, and $T$ hidden layers
consisting of a linear channel-wise transformation $\mathcal{W}_t$,
nonlocal kernel integral operators $\mathcal{K}_t$
and nonlinear activation function $\sigma$
\citep{kovachki2023neural}.
The parameterization of the nonlocal operator $\mathcal{K}$ is where most neural operator architectures vary (cf.~\citep[Section~4]{kovachki2023neural}).

Fourier neural operators (FNOs) \citep{li2020fourier} model $\mathcal{K}$ as a convolution operator, parameterized in the Fourier domain where its application becomes a point-wise multiplication:
\begin{align}
  (\mathcal{K}v)(x)
  = \int_{\mathbb{T}^d} \kappa(x-y)v(y) \dd y
  = \sum_{m \in \mathbb{Z}^d, \|m\|_\infty \le M}
  \widehat{\kappa}_m \, \widehat{v}_m \, e^{i k_m \cdot x},
  \label{eq:fno-layer}
\end{align}
where $\mathbb{T}^d = \mathbb{R}^d / \mathbb{Z}^d$ is the periodic domain and
$\widehat{\kappa}_m$, $\widehat{v}_m$
denote the Fourier coefficients of $\kappa$ and $v$ at the wavevector $k_m = 2\pi m$, for $m \in \mathbb{Z}^d$.
FNOs then parameterize the Fourier coefficients of $\kappa$ as a lookup table of trainable weights
$\widehat{\kappa}_m =: R_m \in \mathbb{C}^{c \times c}$,
where $c$ is the number of channels in the hidden representation $v$,
up to a mode cutoff $\|m\|_\infty \le M$.

A particular strength of FNOs is that they are independent of the resolution at which the input-output pairs are discretized, as long as the grid is fine enough to resolve the retained modes, since finer discretization only adds frequencies beyond the cutoff $M$.
However, FNOs are not independent of the choice of the periodic domain:
Consider a general periodic domain $D = \mathbb{R}^d / L\mathbb{Z}^d$, specified by its lattice matrix $L \in \mathbb{R}^{d \times d}$ whose columns span the domain.
The wavevectors $k_m$ within the Fourier expansion of \cref{eq:fno-layer}
then depend on the lattice $L$ as
\begin{equation}
  k_m = 2\pi L^{-\top} m.
  \label{eq:wavevector}
\end{equation}
Therefore, if one were to apply the same trained FNO on data from two different domains $D$ and $D'$,
the learned weights $R_m$ would be applied at different physical wavevectors $k_m$,
and thus the applied operator $\mathcal{K}$ would differ depending on the domain.
For physical applications where one wants to learn and apply a common operator across varying domains, this is a severe limitation.

\section{Euclidean Fourier neural operators}
\label{sec:efno}
We propose a simple modification of the FNO that removes the implicit dependence on the choice of periodic domain.
Let $\mathcal{K}$ be a translation-invariant linear convolution operator defined in Euclidean space $\mathbb{R}^d$.
Such an operator $\mathcal{K}$ is fully characterized by its \emph{symbol} $\widehat{\kappa} : \mathbb{R}^d \to \mathbb{C}^{c \times c}$, by which it acts in Fourier space as
$\widehat{(\mathcal{K}v)}(k) = \widehat{\kappa}(k)\,\widehat{v}(k)$ for any $k\in\mathbb{R}^d$.
For any given $L$-periodic function $v$,
the action of $\mathcal{K}$ on $v$ can be expressed in real space as a Fourier series
\begin{align}
  (\mathcal{K}v)(x)
  = \sum_{m \in \mathbb{Z}^d} \widehat{\kappa}(k_m) \, \widehat{v}_m \, e^{i k_m \cdot x},
  \qquad \text{with} \quad k_m = 2\pi L^{-\top} m.
  \label{eq:efno-layer}
\end{align}
Instead of learning a lookup table over mode numbers $m$, we propose here to directly parameterize the continuous symbol
$\widehat{\kappa}_\theta : \mathbb{R}^d \to \mathbb{C}^{c \times c}$,
such that it can be evaluated at any wavevector $k$ the domain at hand requires.
We call the resulting operator the \emph{Euclidean Fourier neural operator} (EFNO).

In practice, as in the FNO, we assume that the input-output pairs are discretized on uniform grids such that we can use FFTs to efficiently compute $\widehat{v}_m$ and to reconstruct $(\mathcal{K}v)(x)$ on the grid.
But additionally, this assumption truncates the Fourier series expansion of \cref{eq:efno-layer}:
If $v$ is discretized on a grid $N_1 \times \dots \times N_d$,
the FFT resolves only modes with $|m_j| \le N_j/2$ for $j=1,\dots,d$,
turning the series into a finite sum.
Thus, unlike the FNO, the EFNO does not require a fixed mode cutoff,
but the decay of the symbol governs the convergence with grid resolution.

Parameterizing the continuous symbol $\widehat{\kappa}_\theta(k)$ also enables the incorporation of physical inductive biases, since known properties of the operator $\mathcal{K}$ translate directly into constraints on $\widehat{\kappa}_\theta$:
a smooth real-space kernel $\kappa_\theta(x)$ requires a decaying $\widehat{\kappa}_\theta(k)$;
exponential decay in $\kappa_\theta(x)$ requires smoothness in $\widehat{\kappa}_\theta(k)$;
$E(3)$-equivariance on scalar channels requires an isotropic symbol $\widehat{\kappa}_\theta(k) = \widehat{\kappa}_\theta(\abs{k})$;
and real-valuedness of $\kappa_\theta(x)$ requires Hermitian symmetry $\widehat{\kappa}_\theta(-k) = \overline{\widehat{\kappa}_\theta(k)}$.
See also \cref{sec:kernel-properties} for a more detailed collection of these standard correspondences.

In principle any function approximator can serve as $\widehat{\kappa}_\theta$, including neural networks,
but some model classes make such properties easier to impose than others.
In the experiments below, we consider a linear Gaussian basis model to parameterize an exponentially decaying, smooth, isotropic, and real-valued symbol~$\widehat{\kappa}_\theta$;
see \cref{sec:efno-impl} for details.
Note that the choice of isotropy is motivated by the rotational equivariance of the physical target operators we consider below.
The proposed framework also allows for anisotropic symbols, which may be relevant for other applications.

\section{Experiments}
\label{sec:experiments}

We evaluate the proposed EFNO on two tasks.
First, we consider a simple heat equation whose solution operator is a Gaussian convolution and compare the behavior of a trained FNO and EFNO when evaluated on different periodic domains.
Second, we evaluate the EFNO on a more realistic materials science task
and learn the \emph{exchange-correlation potential} across different crystal structures,
which inherently requires generalization across both grid sizes and domains.
Code for the implementation and the experiments is publicly available on GitHub%
\footnote{\url{https://github.com/epfl-matmat/euclidean-fourier-neural-operators}}.

\subsection{Transferring a learned convolution across equivalent periodic cells}
\label{sec:exp1}

\begin{figure}[t]
  \centering
  \includegraphics[width=\textwidth]{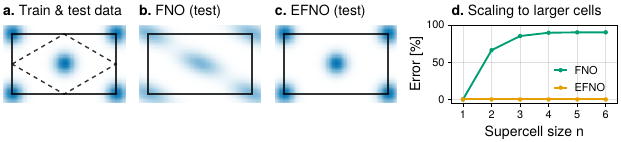}
  \caption{
    \textbf{Domain transfer test for a learned convolution.}
    $\textbf{a}.$ A periodic array of Gaussian sources, smoothed by the heat kernel.
    Both models are fitted in the small rhombic cell (dashed),
    and evaluated on the equivalent rectangular cell (solid).
    The FNO ($\textbf{b}.$) distorts visibly, while the EFNO ($\textbf{c.}$) transfers perfectly.
    \textbf{d.}~Relative $L_2$ error of the FNO and EFNO on $n\times n$ supercells of the fitting cell.
    The FNO error grows with the size of the supercell, while the EFNO error remains constant.
  }
  \label{fig:exp1}
\end{figure}

This experiment evaluates the abilities of FNO and EFNO to transfer a learned convolution across periodic domains that can represent the same physical system but have different shapes and sizes.
To this end, consider the solution operator of the heat equation $\partial_t u = \Delta u$ with periodic boundary conditions,
$\mathcal{G}^{(t)}\colon u(0,\cdot) \mapsto u(t,\cdot)$,
which is a convolution with symbol $\widehat{\kappa}^{(t)}(k) = \operatorname{exp}(-t\abs{k}^2)$.
We fit single FNO and EFNO spectral convolution layers to the exact symbol for a fixed $t = 0.006$ on a small rhombic cell,
and then evaluate the fitted operators on an equivalent rectangular cell of twice the area and on $n \times n$ supercells of the fitting cell.
Details on the exact experimental setup and the model architectures are provided in
\cref{sec:exp1-details}.

\Cref{fig:exp1} shows the results.
The FNO shows precisely the domain dependence failure mode described in \cref{sec:fno} and does not transfer the learned convolution from the rhombic cell to the rectangular cell,
whereas the EFNO transfers perfectly.
Similarly for the $n \times n$ supercells, the relative $L_2$ error of the FNO grows with the size of the supercell to $\sim\!90\%$.
The EFNO error remains constant below $10^{-5}\%$, which demonstrates its capabilities to transfer a learned operator across periodic domains.

\subsection{Learning exchange-correlation potentials across crystal structures}

We now apply the EFNO to a materials science task that requires generalization across both resolutions and domains.
The de-facto standard method for simulating the electronic structure of atoms, molecules, and solids is density-functional theory (DFT),
but in practice the accuracy of DFT is limited by the accuracy of the \emph{exchange-correlation}~(XC) functional.
Various XC approximations have been proposed, ranging from cheap models like the PBE functional \citep{perdew1996generalized}, to approaches like the random phase approximation (RPA)~\citep{Furche2001, Fuchs2002}
that are more accurate but are prohibitively expensive for routine use.
We consider here the task of learning an operator that maps PBE electron densities to RPA exchange-correlation potentials.

\begin{figure}[t]
  \centering
  \includegraphics[width=\textwidth]{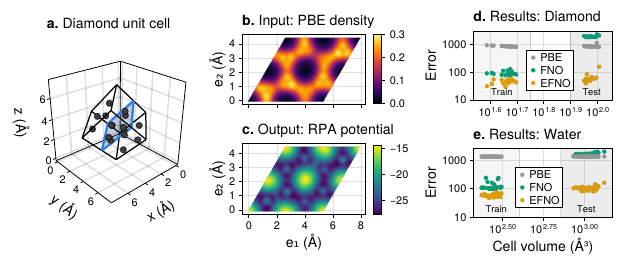}
  \caption{
    \textbf{Overview and results of the exchange-correlation potential learning task.}
    \textbf{a.} Example system, showing a diamond unit cell with 16 atoms.
    \textbf{b.--c.} PBE electron density and RPA exchange-correlation potential on the slice plane highlighted in (\textbf{a.}); these are the inputs and outputs for the operator learning task.
    \textbf{d.--e.} Error (WRMSE) against cell volume for both tasks.
    The FNO extrapolates poorly from the smaller training cells to the larger test cells; the EFNO generalizes well and achieves errors that are an order of magnitude smaller than the PBE baseline.
  }
  \label{fig:xc-experiment}
\end{figure}

We consider two subsets of the ML-RPA dataset by \citet{riemelmoser2023} of PBE densities and RPA potentials:
\emph{diamond} consisting of 20 8-atom structures and 20 16-atom structures, and \emph{liquid water} consisting of 32 structures of 8 water molecules as well as 39 structures of 31 or 32 water molecules.
A visualization of a diamond system and its density and potential is shown in \cref{fig:xc-experiment}.
For each subset, we train operators on the smaller structures and evaluate them on the larger ones.

We use the same EFNO architecture for both tasks, consisting of a lifting layer, 2 EFNO layers, and a projection layer, with 8 channels and 16 basis functions.
We additionally train a standard FNO with 2 layers, 8 channels, and 4 modes as a baseline, that we apply across cell sizes without modification.
Both model architectures were chosen by minimizing error on a validation split of the training data.
We train both models on a gauge-invariant density-weighted mean squared error loss,
and evaluate the trained models with the square-root of this metric (WRMSE).
More details on the architecture and the training and evaluation procedure are provided in \cref{sec:exp-details:arch}.

\begin{wraptable}{r}{7.5cm}
  \vspace{-1.5\intextsep}
  \centering
  \caption{Density-weighted root mean square error (WRMSE); median value within each split.}
  \label{tab:xc-results}
  \begin{tabular}{lccc ccc}
    \toprule
    & \multicolumn{2}{c}{Diamond} & \multicolumn{2}{c}{Water} \\
    Model & Train & Test & Train & Test \\
    \midrule
    PBE  & 826.67 & 825.17 & 1350.33 & 1348.37 \\
    FNO  & 81.96 & 1815.42 & 108.03 & 1675.35 \\
    EFNO & \textbf{47.88} & \textbf{56.11} & \textbf{57.95} & \textbf{103.54} \\
    \bottomrule
  \end{tabular}
  \vspace{-1.0\intextsep}
\end{wraptable}
\Cref{tab:xc-results}
and \cref{fig:xc-experiment}
show the main results.
Both the EFNO and FNO are able to fit the training data well and return potentials that are significantly closer to RPA than the PBE baseline.
However, the FNO fails to generalize to the larger test structures.
The EFNO also shows an increase in test error, but still outperforms PBE by an order of magnitude throughout all instances.
More visualizations of the predicted potentials for a few example structures can be found in \cref{sec:exp-details:slices}.
Altogether, the results demonstrate that the EFNO is able to learn an operator that generalizes across both grid sizes and domains, showing its potential for applications in materials science and chemistry.

\section{Conclusion}
\label{sec:conclusion}
We proposed Euclidean Fourier neural operators (EFNOs).
By parameterizing the Fourier symbol as a continuous function of the physical wavevector, EFNOs remove the implicit domain dependence of FNOs and learn operators that can be applied consistently across domains of varying shape and size.
This has been demonstrated in experiments on a heat equation and on a materials science task of learning exchange-correlation potentials across crystal structures.
Natural extensions of this work include more expressive symbol parameterizations beyond isotropic Gaussian models, comparison to additional baselines, and evaluation on complex PDEs from other fields,
to further establish EFNOs as a practical approach to operator learning in the physical sciences.

\begin{ack}
This research was supported by the Swiss National Science Foundation (SNSF,
Grant No.~10002757) as well as the NCCR MARVEL, a National Centre of Competence
in Research, funded by the SNSF (Grant No.~205602).
We express our gratitude to Stefan Riemelmoser for helpful discussions
and to Bruno Ploumhans for feedback on the manuscript.
\end{ack}

\bibliographystyle{plainnat}
\bibliography{references}

\appendix

\section{Related work}
\label{sec:related}

\paragraph{Neural operators on varying geometries.}
Fourier neural operators \citep{li2020fourier} apply learned per-mode weight matrices to truncated Fourier coefficients, within the general operator-learning framework of \citet{kovachki2023neural}.
Several variants extend them to irregular or varying geometry by deforming the domain to a regular grid (Geo-FNO, \citealp{li2022geo}), mapping it to a fixed reference domain (DNO, \citealp{zhao2024diffeomorphism}), encoding it with a smoothed characteristic function (DAFNO, \citealp{liu2023dafno}), replacing the FFT with a non-uniform transform \citep{lingsch2024beyond}, or supplying the domain geometry as an extra input channel \citep{oguadimma2026geometry}; in each case the spectral weights stay indexed by mode number, so transferring across physical domains still requires re-fitting or re-learning them.
Non-spectral operators such as DeepONet \citep{lu2021deeponet} and graph neural operators \citep{li2020graph} handle varying geometry natively through query coordinates or message passing, respectively, but forgo the FFT and its efficiency.

\paragraph{Learned continuous spectral symbols.}
A closer line of work replaces the discrete weight table with a function of the frequency variable.
AM-FNO \citep{xiao2024amortized} and SirenFNO \citep{shi2026sirenfno} learn a network over the mode number (or a grid-normalized version of it), covering all modes with a fixed parameter budget.
PDNO \citep{shin2024pdno} and KANO \citep{lee2026kano} instead read the weight as a pseudo-differential symbol that also depends on position, for variable-coefficient problems.
PDNO formulates its symbol over $\xi \in \mathbb{R}^d$ but restricts the symbol network to integer modes $\xi \in \mathbb{Z}^d$ (their Appendix~B2), a choice they motivate by continuity of the operator.
In all of these the frequency argument is the mode number, so the represented operator depends on the domain, and the weights are trained and evaluated at a single physical scale.
Our proposed EFNO layer keeps a position-independent convolution but instead takes the physical wavevector as the argument (\cref{tab:relwork}).
A similar approach has been concurrently proposed by \citet{khan2026ksfno} who also target transfer across differently sized domains, with radially factorized spectral filters as functions of physical reciprocal-space radius.

\begin{table}[h]
  \centering
  \caption{Parameterization of the spectral weight across FNO variants. FNO stores a discrete table over modes; the others learn it as a function. Its argument is the mode number, with PDNO and KANO adding the position (a pseudo-differential symbol); EFNO alone uses the physical wavevector, the quantity that changes with the domain.}
  \label{tab:relwork}
  \begin{tabular}{lll}
    \toprule
    Method & Spectral weight & Kernel argument \\
    \midrule
    FNO \citep{li2020fourier}        & lookup table     & mode number \\
    AM-FNO \citep{xiao2024amortized} & learned function & mode number \\
    SirenFNO \citep{shi2026sirenfno} & learned function & mode number \\
    PDNO \citep{shin2024pdno}        & learned function & mode number \& position \\
    KANO \citep{lee2026kano}         & learned function & mode number \& position \\
    EFNO (ours)                      & learned function & physical wavevector \\
    \bottomrule
  \end{tabular}
\end{table}

\paragraph{Applications to electronic structure.}
Closest to our application are neural operators that predict electronic-structure fields.
V2Rho-FNO \citep{jin2026v2rho} maps the external potential to the electron density for molecules in a fixed cubic box, and GPWNO \citep{kim2024gpwno} predicts electron densities across crystals using a combined orbital and plane-wave basis; in both cases the spectral weights are tied to a single domain or bypass the Fourier-multiplier structure.
\citet{khan2026ksfno} use their concurrently developed domain-invariant FNO to learn the Kohn--Sham map from potential to density across atomistic systems of varying size.
Our experiment learns a map from PBE densities to RPA exchange-correlation potentials, on structures whose lattices differ from sample to sample.
Representing an operator by a continuous Fourier symbol of the physical wavevector is standard practice in plane-wave electronic-structure methods, for example for the Coulomb kernel or the nonlocal van der Waals kernel \citep{romanperez2009}.

\section{Kernel properties}
\label{sec:kernel-properties}

\Cref{tab:kernel-properties} collects some
standard correspondences from Fourier analysis
connecting the real-space convolution kernel and its Fourier transform.

\begin{table}[h]
  \centering
  \caption{Correspondence between real-space and spectral properties of the kernel.
    Each row is an equivalence.}
  \label{tab:kernel-properties}
  \begin{tabular}{ll}
    \toprule
    Real-space kernel $\kappa$ & Spectral kernel $\widehat{\kappa}$ \\
    \midrule
    \emph{Real-valuedness} & \emph{Hermitian symmetry}\\
    $\kappa$ real-valued & $\widehat{\kappa}(-k) = \overline{\widehat{\kappa}(k)}$ \\
    \cmidrule(lr){1-2}
    \emph{Smoothness} & \emph{Decay at large $\abs{k}$} \\
    $\kappa$ has $s$ derivatives in $L^2$ & $\abs{k}^s\, \widehat{\kappa} \in L^2$ \\
    \cmidrule(lr){1-2}
    \emph{Range} & \emph{Smoothness of $\widehat{\kappa}$} \\
    $\kappa$ supported in $\abs{x} \le R$ & $\widehat{\kappa}$ entire, $\abs{\widehat{\kappa}(k)} \lesssim e^{R \abs{\operatorname{Im} k}}$ \\
    \cmidrule(lr){1-2}
    \emph{Isotropy} & \emph{Isotropy} \\
    $\kappa(x) = \kappa(\abs{x})$ & $\widehat{\kappa}(k) = \widehat{\kappa}(\abs{k})$ \\
    \bottomrule
  \end{tabular}
\end{table}

\section{Gaussian basis parameterization of the EFNO symbol}
\label{sec:efno-impl}

We parameterize the symbol $\widehat{\kappa}_\theta$ as a linear combination of Gaussian basis functions in $\lvert k \rvert^2$,
\begin{equation}
  \widehat{\kappa}_\theta(k) = \sum_{j=1}^{n_\text{basis}} \phi_j\!\left( \lvert k \rvert^2 \right) \, w_j,
  \qquad
  \phi_j\!\left( \lvert k \rvert^2 \right) = \exp\!\left( \frac{ -(\lvert k \rvert^2 - \mu_j)^2 }{ 2\sigma^2 } \right),
  \label{eq:gaussian-basis}
\end{equation}
with
$n_\text{basis}$ centers $\mu_j$ placed uniformly on $[0, k_\text{max}^2]$ in $\lvert k \rvert^2$,
a shared width $\sigma = k_\text{max}^2/n_\text{basis}$, and
trainable real coefficients $w_j \in \mathbb{R}^{c \times c}$.
Since the basis is Gaussian in $\lvert k \rvert^2$, the symbol $\widehat{\kappa}_\theta(k)$ is a smooth function of $k$, decays as $\lvert k \rvert \to \infty$, is isotropic as it depends only on $\lvert k \rvert$, and is real-valued by construction;
see \cref{sec:kernel-properties} for the correspondences between these properties and those of the real-space kernel.
Our choice of isotropy is adapted to the underlying symmetry of the
learning task in the two experiments considered here. 
More generally, the symbol could depend on the direction of $k$ as well, 
for example when learning anisotropic operators or to act on vectorial or tensorial features.

\section{Additional experiment details and results}
\label{sec:exp-details}

\subsection{Experiment 1: Gaussian smoothing across periodic cells}
\label{sec:exp1-details}

\paragraph{Operator and cells.}
The test operator is the solution operator of the heat equation
$\partial_t u = \Delta u$ with periodic boundary conditions,
$\mathcal{G}^{(t)}\colon u(0,\cdot)\mapsto u(t,\cdot)$ at $t = 0.006$,
a convolution with the known symbol
$\widehat{\kappa}^{(t)}(k) = \exp(-t\,\lvert k\rvert^2)$.
Both models are fitted on a rhombic cell with lattice
\begin{equation}
  L_\text{fit} = \begin{pmatrix} \sqrt{3}/2 & \sqrt{3}/2 \\ 1/2 & -1/2 \end{pmatrix},
\end{equation}
sampled on a $48\times 48$ grid.
Models are then evaluated on two different sets of test cells:
\begin{enumerate}[leftmargin=*,label=(\roman*)]
\item a rectangular cell of twice the area,
  $L_\text{rect} = \begin{pmatrix} \sqrt{3} & 0 \\ 0 & 1 \end{pmatrix}$,
  on an $84\times 48$ grid;
\item $n\times n$ supercells with lattice $L_n = n L_\text{fit}$, for $n = 1,\dots,6$,
  on $48n\times 48n$ grids, holding the resolution per unit length constant.
\end{enumerate}
With this choice, any field that is periodic on $L_\text{fit}$ is also periodic on $L_\text{rect}$ and on $L_n$ for all $n$.

\paragraph{Models and fitting.}
Both evaluated models consist only of a single spectral convolution layer so that the only difference is the parameterization of the symbol.
The FNO keeps $M = 12$ modes per dimension ($325$ weights);
The EFNO uses the Gaussian basis of \cref{eq:gaussian-basis} with $n_\text{basis} = 25$ coefficients, centers uniform on $[0, k_\text{max}^2]$ with $k_\text{max}^2 = -\log(10^{-6})/t$ and width $\sigma = k_\text{max}^2/n_\text{basis}$.
Both are fitted in closed form to the exact symbol on the fitting cell's wavevector grid, with no input-output data and no iterative optimization:
the FNO sets $R_m = \widehat{\kappa}(k_m)$ directly,
the EFNO solves a least-squares problem with a small Tikhonov term ($10^{-10}\,\mathrm{tr}(\Phi^\top \Phi)/n_\text{basis}\, I$) for numerical conditioning.

\paragraph{Input field and evaluation.}
The probe is a periodic array of Gaussian sources (width $0.03$, one per lattice site, summed over periodic images).
As it has the periodicity of the rhombic fitting cell, it is exactly representable on every supercell.
Errors are relative $L_2$ against the exact smoothed field, in percent.

\paragraph{Results.}
\Cref{tab:exp1-results} reports the error on the fitting cell, the rectangular supercell, and the $n \times n$ supercells.
\Cref{fig:exp1} visualizes the rectangular-cell transfer and the supercell sweep.
On the fitting cell, both models are near-exact.
On both the rectangular test cell and the $n \times n$ supercells, the FNO error becomes large while the EFNO error remains constant at the fitting residual.

\begin{table}[h]
  \centering
  \small
  \caption{
    Relative $L_2$ error (\%) of the FNO and EFNO.
    The FNO is near-exact on the fitting cell but fails to transfer;
    the EFNO stays at its basis-approximation residual on every cell.
  }
  \label{tab:exp1-results}
  \begin{tabular}{lccccccc}
    \toprule
    & \multicolumn{2}{c}{Single cell} & \multicolumn{5}{c}{$n\times n$ supercell} \\
    \cmidrule(lr){2-3} \cmidrule(lr){4-8}
    Model & Fitting & Rectangular & $n=2$ & $n=3$ & $n=4$ & $n=5$ & $n=6$ \\
    \midrule
    FNO  & $\sim\!0$ & $49.70$ & $66.09$ & $85.16$ & $89.60$ & $90.12$ & $90.16$ \\
    EFNO & $3.5{\times}10^{-6}$ & $3.5{\times}10^{-6}$ & $3.5{\times}10^{-6}$ & $3.5{\times}10^{-6}$ & $3.5{\times}10^{-6}$ & $3.5{\times}10^{-6}$ & $3.5{\times}10^{-6}$ \\
    \bottomrule
  \end{tabular}
\end{table}

\subsection{Experiment 2: Exchange-correlation potential learning}
\label{sec:exp2-details}

\subsubsection{Architecture and training}
\label{sec:exp-details:arch}

\paragraph{Dataset and splits.}
We use the ML-RPA dataset \citep{riemelmoser2023}, consisting of 189 crystal structures
spanning small molecules, diamond bulk systems, surfaces, and liquid water.
We consider two subsets:
the \emph{diamond} subset, consisting of 20 eight-atom structures for training and validation, and 20 sixteen-atom structures for testing;
and the \emph{liquid-water} subset, consisting of 32 structures of eight water molecules for training and validation, and 39 structures of 31 or 32 water molecules for testing.
In both cases, we train on the smaller structures and test on the larger ones, with an 80/20 train/validation split.
The training and validation structures are Fourier-resampled to a common grid
($48^3$ for diamond, $80^3$ for water)
to enable efficient batching during training;
the test structures are kept on their original grids, so that the evaluation reflects generalization across both grid sizes and domains.

\paragraph{Loss and evaluation metric.}
Following \citet{riemelmoser2023}, we train the models on a gauge-invariant density-weighted mean squared error loss,
\begin{equation}
  \mathcal{L}
  = \frac{1}{N_e} \int_{\Omega} \rho(r) \,\bigl(\tilde{v}_{\text{xc}}^{\text{pred}}(r) - \tilde{v}_{\text{xc}}^{\text{ref}}(r)\bigr)^2\dd r,
  \label{eq:xc-loss}
\end{equation}
where $N_e = \int_{\Omega} \rho(r) \dd r$ is the number of electrons in the cell $\Omega$, and $\tilde{v}_{\text{xc}} = v_{\text{xc}} - \tfrac{1}{N_e}\int_{\Omega} \rho(r) \, v_{\text{xc}}(r) \dd r$ subtracts the density-weighted mean to remove the constant gauge freedom of $v_\text{xc}$.
All integrals are discretized over the uniform real-space grid.
We train on the mean of $\mathcal{L}$ over the training structures, and report its square root (WRMSE) in meV/$e^-$.

\paragraph{Model and training.}
Each model consists of a lifting layer that maps the scalar input to $c$ channels, two (E)FNO blocks, and a projection layer that maps back to a single channel.
As in the standard FNO, each block is a sum of a spectral convolution and a pointwise linear channel-mixing layer, followed by a \texttt{gelu} activation.
We selected the architectures by a hyperparameter search over the number of channels $c$, the number of (E)FNO layers, and the number of basis functions or modes per layer, minimizing the validation WRMSE.
The resulting models both use $c = 8$ channels, two (E)FNO layers, a lifting and a projection layer.
The EFNO uses 16 basis functions per layer, spread uniformly in $[0, k_\text{max}^2]$ with $k_\text{max}^2 = 40$ bohr$^{-2}$, chosen to roughly match the orbital plane-wave cutoff of the ML-RPA dataset ($E_\text{cut} = 600$~eV).
The FNO baseline uses 4 modes per layer.
Both models are trained with Adam at a constant learning rate of $10^{-2}$, full-batch over all training structures, for a fixed number of steps, and we select the checkpoint with the lowest validation WRMSE.

\subsubsection{Visualizations of the predicted potentials}
\label{sec:exp-details:slices}

\begin{figure}[t]
  \centering
  \includegraphics[width=\textwidth]{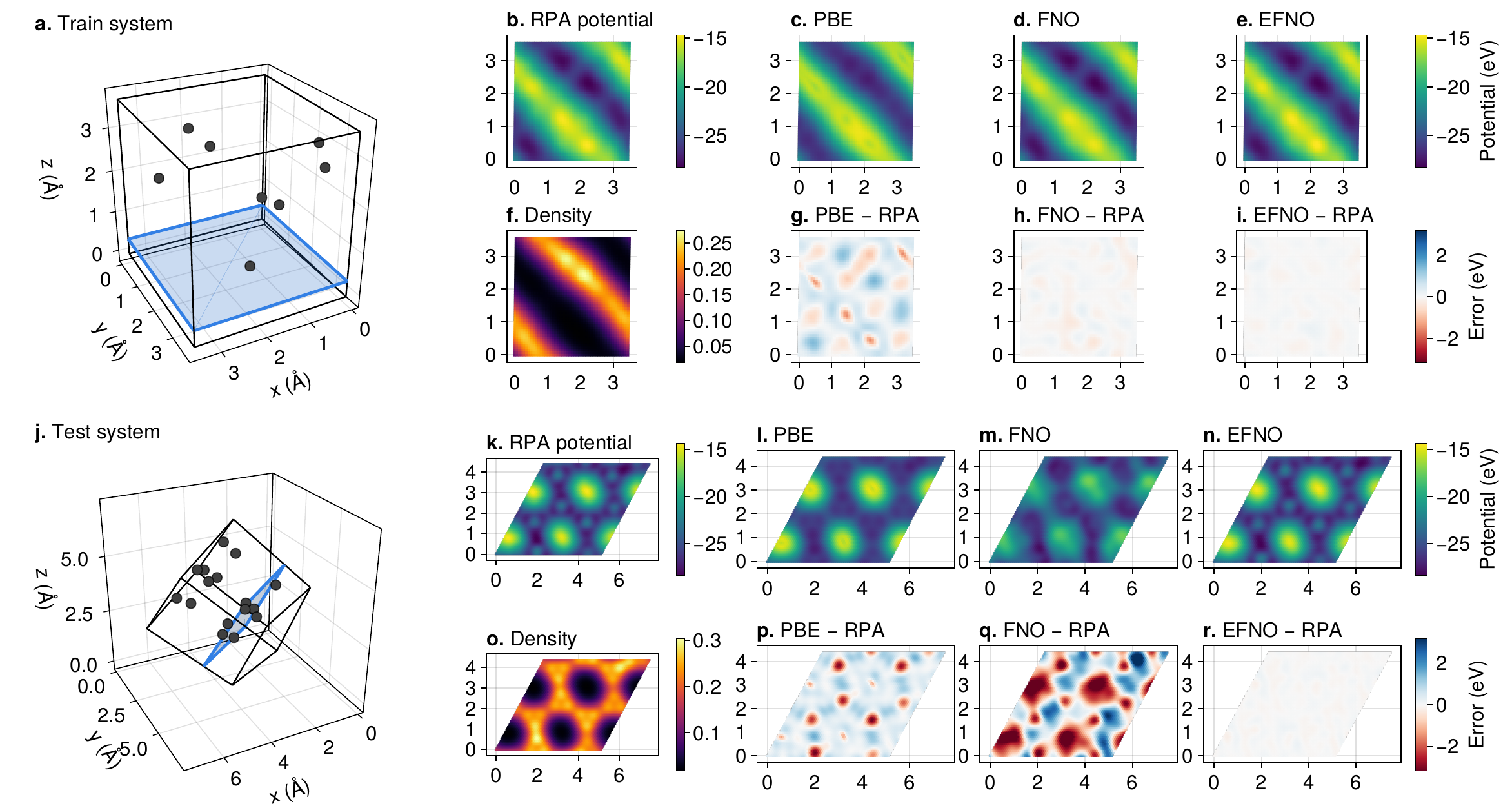}
  \caption{
    \textbf{Predicted potentials for a representative diamond train and test structure.}
    \textbf{a.}~Unit cell of a training system, showing the plane along which the density and potential are visualized.
    \textbf{b.}~RPA reference potential.
    \textbf{c.--e.}~PBE, FNO, and EFNO predictions.
    \textbf{f.}~Input electron density (PBE).
    \textbf{g.--i.}~Difference of each prediction to the reference.
    \textbf{j.--r.}~Corresponding panels showing the cell, potentials, density, and errors for a test system.
    On the training system, both the FNO and the EFNO are able to reproduce the RPA potential well and show small errors (\textbf{h.--i.}).
    On the test system, the FNO shows substantial errors (\textbf{q.}) that are larger than the PBE baseline (\textbf{p.}), while the EFNO stays close to the reference (\textbf{r.}).
  }
  \label{fig:xc-slices-diamond}
\end{figure}

\begin{figure}[t]
  \centering
  \includegraphics[width=\textwidth]{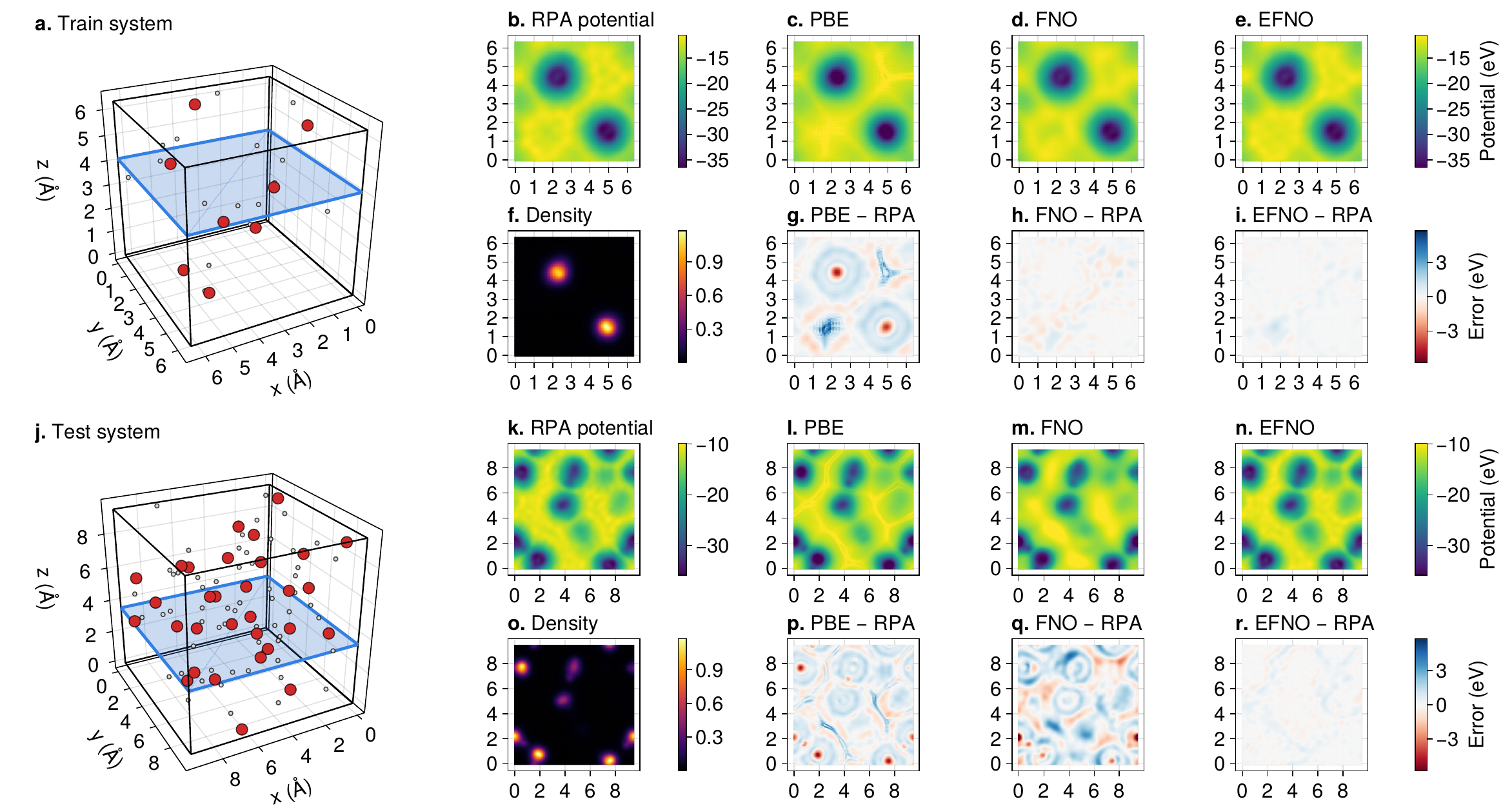}
  \caption{
    \textbf{Predicted potentials for a representative liquid-water train and test structure.}
    \textbf{a.}~Unit cell of a training system, showing the plane along which the density and potential are visualized.
    \textbf{b.}~RPA reference potential.
    \textbf{c.--e.}~PBE, FNO, and EFNO predictions.
    \textbf{f.}~Input electron density (PBE).
    \textbf{g.--i.}~Difference of each prediction to the reference.
    \textbf{j.--r.}~Corresponding panels showing the cell, potentials, density, and errors for a test system.
    On the training system, both the FNO and the EFNO are able to reproduce the RPA potential well and show small errors (\textbf{h.--i.}).
    On the test system, the FNO shows substantial errors (\textbf{q.}) that are larger than the PBE baseline (\textbf{p.}), while the EFNO stays close to the reference (\textbf{r.}).
  }
  \label{fig:xc-slices-water}
\end{figure}

\Cref{fig:xc-slices-diamond,fig:xc-slices-water} show the predicted RPA exchange-correlation potentials for one representative training and one representative test structure from each task (diamond and liquid water), with the PBE electron density as input.
On the training structures both models reproduce the RPA potential;
on the test structures the FNO error grows substantially while the EFNO stays close to the reference, which is consistent with the aggregated results of \cref{tab:xc-results}.

\clearpage

\end{document}